\documentclass[11pt,letterpaper]{article}
\usepackage{arxiv}

\usepackage{hyperref}
\usepackage{url}
\usepackage{amsmath,amssymb,amsthm,amsfonts}
\usepackage{algorithm}
\usepackage{graphicx}
\usepackage{subcaption}
\usepackage[inline]{enumitem}
 \usepackage{natbib}
\usepackage{enumitem}
\usepackage[noend]{algpseudocode}

\title{Rethinking Optimization: A Systems-Based Approach to Social Externalities}
\author{Pegah Nokhiz\\
  Cornell University, Cornell Tech\\
  \texttt{pegah.nokhiz@gmail.com} \\
  \And
   Aravinda Kanchana Ruwanpathirana \\
  National University of Singapore\\
\texttt{kanchana.ruwanpathirana@gmail.com}\\
  \And
  Helen Nissenbaum  \\
  Cornell Tech\\
  \texttt{hn288@cornell.edu} \\}

\begin{document}



\date{}
\maketitle
\begin{abstract}
Optimization is widely used for decision making across various domains, valued for its ability to improve efficiency. However, poor implementation practices can lead to unintended consequences, particularly in socioeconomic contexts where externalities (costs or benefits to third parties outside the optimization process) are significant. To propose solutions, it is crucial to first \textit{characterize} involved stakeholders, their goals, and the types of subpar practices causing unforeseen outcomes. This task is complex because affected stakeholders often fall outside the direct focus of optimization processes. Also, incorporating these externalities into optimization requires going beyond traditional economic frameworks, which often focus on describing externalities but fail to address their normative implications or interconnected nature, and feedback loops. 

This paper suggests a framework that combines systems thinking with the economic concept of externalities to tackle these challenges. This approach aims to characterize\textit{ what went wrong, who was affected, and how (or where) to include them in the optimization process.} Economic externalities, along with their established quantification methods, assist in identifying ``who was affected and how" through stakeholder characterization. Meanwhile, systems thinking (an analytical approach to comprehending relationships in complex systems) provides a holistic, normative perspective. Systems thinking contributes to an understanding of interconnections among externalities, feedback loops, and determining ``when" to incorporate them in the optimization. Together, these approaches create a comprehensive framework for addressing optimization's unintended consequences, balancing descriptive accuracy with normative objectives. Using this, we examine three common types of subpar practices: ignorance, error, and prioritization of short-term goals.

\end{abstract}

\section{Introduction}
Optimization is a key tool for decision-making in various domains, including industrial engineering, urban planning, social systems, and business, enabling efficient resource allocation and driving advancements in areas like artificial intelligence and machine learning \cite{bottou2010large}. However, optimization simplifies the complexity of real-world systems by focusing on specific variables or objectives, often neglecting the broader social, environmental, or systemic effects of decisions. While this approach can provide clear and actionable solutions, it results in poor practices in applying optimization that frequently lead to unintended consequences \cite{berk2021fairness,Buolamwini2018,McMillan2011}. Optimization, often carried out as a descriptive process, underpins the creation of many computational tools like machine learning models. However, it also leads to unintended outcomes that inherently carry normative inquiries \cite{laufer2023optimization} that bring attention to the impact of actions on involved parties, in particular, within complex and emergent socioeconomic systems.

As a community focused on ML and AI research, we aim to address these unintended consequences. However, before we can propose effective solutions for such unintended consequences, it is essential to recognize that simply labeling all undesired aftereffects as unintended consequences is problematic. This is because: (1) it is overly broad, failing to account for the nuances and specific differences between cases that require tailored approaches, and (2) it implies that all problematic outcomes are uniformly distributed or {objectively} problematic, which may not always be the case.

To effectively resolve unintended consequences, it is essential to \emph{characterize}  their types (e.g., unexpected benefits or drawbacks \cite{merton2003obituary}), causes (e.g., optimizers' ignorance or error), and all affected parties. Only with these characterizations established, targeted solutions can be developed. For example, in a company, optimizing water cooler waiting times by assigning time slots may enhance efficiency but unintentionally disrupt informal social interactions \cite{Weeks07Water, Orn22Covid, Lynne17Water}, diminishing collaboration, ultimately harming both employees and the organization. To address this, it is crucial to identify \emph{who is affected}, \emph{how}, and \emph{when} communal values should be integrated into the optimization. This characterization would also help pinpoint the underlying subpar practice in deploying optimization (like ignorance, mistakes, or a focus on short-term gains instead of long-term goals \cite{merton2003obituary, merton1936unanticipated}) that caused unforeseen outcomes.

However, characterizing optimization's unintended consequences is challenging due to the difficulty in determining their extent in social contexts. Traditional models focus on specific objectives, often overlooking indirect effects like broader social values or stakeholder impacts. For example, benefits such as social interactions in the workplace may be excluded from the water cooler optimization. Identifying all stakeholders (those affected by decisions) and quantifying the impacts on them is critical. This requires both \emph{descriptive} (recognizing stakeholders and their goals) and \emph{normative} considerations (addressing the omission of their interests) for characterizations and developing solutions.

\paragraph{Externalities and Internalization.} To address these challenges, we suggest characterizing unintended consequences of optimization in complex socioeconomic systems as externalities, i.e., outcomes that affect stakeholders not directly involved in the decision-making process. While not widely discussed in ML and AI communities in optimization-based computations, externalities are well-established phenomena in economics, defined as costs or benefits incurred by third parties due to an activity \cite{coase1960problem, pigou1920economics}. Solutions to externalities often involve internalization methods such as cost-benefit analysis and Pigouvian tax \cite{pigou1920economics}, which adjust incentives to account for societal costs or benefits. For instance, carbon taxes aim to internalize the social costs of emissions. This procedure aligns individual incentives with broader societal welfare and it ensures that social and communal values are preserved alongside efficiency \cite{Fleurbaey2023}.

The externality framework as a descriptive tool is beneficial in two key ways: (1) It enables the characterization of unintended consequences by distinguishing between positive externalities (benefits to third parties) and negative externalities (costs to third parties). This involves identifying stakeholders and their goals which also helps detect the subpar practice causing the externality. (2) It provides well-established, quantifiable solutions for addressing subpar practices by utilizing commonly used methods like cost-benefit analysis, social welfare assessments, taxes, subsidies, and policy interventions to measure and internalize externalities based on their nature \cite{pigou1920economics, coase1960problem, Arrow1951ARRIVA, Boardman2018, Weitzman1974}.

However, open complex systems like societies have emergent behaviors, feedback loops, and interconnections among different parts of the system. These significantly influence social externalities \cite{Fleurbaey2023}. Feedback loops (positive or negative) can amplify the effects of social externalities, making them far more significant than initially perceived \cite{Fleurbaey2023, Satz2010, Kanbur2004, Stern2014}. For example, negative feedback loops, such as declining morale or eroding trust due to poorly designed water cooler interventions, can
spiral into long-term company dysfunction. These dynamics make the management of social externalities (e.g., determining where or when to internalize them) difficult. Economic methods as descriptive toolkits are precise in quantification but they are too narrow in scope to address these challenges.

\paragraph{Systems Thinking of Optimization.} To overcome these challenges, we propose incorporating systems thinking as an added layer of perspective. Systems thinking is a problem-solving and analytical approach that emphasizes understanding the interactions and interdependencies among components of a system within the context of a broader whole \cite{meadows2008thinking, sterman2000business}. This holistic mindset to decision-making emphasizes interconnections, feedback loops (where targeted actions could amplify positive effects or mitigate negative feedback loops), and emergent properties within complex systems \cite{meadows2008thinking, dorfler2024towards}. Unlike traditional optimization that isolates variables, systems thinking considers hidden dependencies and long-term ripple effects. Accordingly, applying a systems thinking mindset in practice, involves systems theoretic mathematical architectures with the same purpose. For example, a layered systems theoretic mathematical architecture~\cite{sarjoughian2001,voros2005,jensen1970linear,Smith1973Hierarchical,Zhu2015HierarchicalAO,matni2024theory,empoweredFive} would help with thinking about ``when or where should be"  the level at which externalities are addressed: lower levels focus on localized stakeholder goals. Higher levels analyze broader overarching goals.

\paragraph{Our Position: Why Neither Concept Alone is Sufficient?} Neither systems thinking nor economic externalities alone can fully address the complexity of identifying, quantifying, and internalizing unintended consequences in optimization: 


\begin{itemize}
    \item \textbf{Externalities for stakeholders:} Who is affected? How?
    \item \textbf{Systems thinking for interconnected dynamics:} When or where do we internalize externalities? How do these impacts propagate, and what secondary or tertiary effects emerge over time? In the water cooler case, externalities identify the direct loss of collaboration (a cost). Systems thinking shows the ripples through the company affecting long-term performance and when to internalize the cost.
\end{itemize}

This paper advocates for a combination of both perspectives, in particular, in widely descriptive optimization-based computations in AI/ML with numerous unforeseen outcomes \cite{laufer2023optimization,berk2021fairness,Buolamwini2018}. Combining externalities with systems thinking creates a powerful framework to address unintended consequences. Externalities provide the descriptive foundation, highlighting where optimizations (as another descriptive tool) fall short. Systems thinking adds normative reflections of \emph{what should be}, ensuring that solutions align with societal values and account for complexity. Together, they address the issues noted earlier: how to identify, quantify, and internalize unintended outcomes. Thus, in this \textit{position} paper, we present three key messages:


\begin{itemize}
\item To effectively resolve the unintended consequences of optimization, it is essential to move beyond treating all unforeseen outcomes as a single category. Instead, we must carefully analyze what went wrong, identify who was impacted, and determine how and where to incorporate these considerations into the optimization process.
\item We propose externalities as a descriptive tool to identify impacted stakeholders, classify effects (costs or benefits), and use proper quantification methods for internalization.
\item We propose systems thinking and a layered system-theoretic model for a normative perspective on when to internalize externalities. This also accounts for local and broader goals, interconnections, and feedback loops within complex systems.
\end{itemize}
\section{Related Work}
\label{sec:related-work}
\textbf{Economic Externalities.} Economic Externalities are unintended costs/benefits experienced by third parties outside market transactions. They arise from unaccounted spillover effects within standard markets, such as pollution (negative) or societal benefits of scientific research (positive). These externalities highlight the misalignment between private incentives and social welfare \cite{pigou1920economics, coase1960problem}.

To address externalities, several methods are developed for internalization. Pigouvian taxes and subsidies correct negative and positive externalities by aligning private and social costs \cite{pigou1920economics}. Social welfare functions incorporate externalities into utility functions to measure societal well-being \cite{Arrow1951ARRIVA}. Cost-benefit analysis (CBA) compares costs and benefits of externalities to guide decision-making \cite{Boardman2018}. Regulatory mechanisms like emission caps and Cap-and-Trade systems limit harmful activities \cite{Weitzman1974}. Coasian bargaining offers negotiation-based solutions when property rights are clearly defined and transaction costs are low \cite{coase1960problem}. Also, recent research has extended externality analysis to social and environmental contexts, including urban planning, education, and algorithmic systems \cite{Arrow1969, Fleurbaey2023}. However, traditional methods face limitations, such as undermining intrinsic motivations in social interactions, prioritizing efficiency over equity, and failing to account for long-term feedback loops and emergent behaviors \cite{Satz2010, Kanbur2004, Stern2014}.

\vspace{0.15cm}

\noindent\textbf{Optimization's Unintended Consequences.} Optimization, while powerful, frequently simplifies complex real-world scenarios, leading to unintended consequences. Fairness-aware machine learning models, for example, can inadvertently create inequalities among subgroups \cite{berk2021fairness, gupta2019equalizing, stinar2022algorithmic, nokhiz2021precarity, nokhiz2024agent, nokhiz2025counting, nokhiz2024modeling}. Algorithms designed to maximize overall success often misclassify underprivileged groups, prioritizing the majority \cite{Buolamwini2018, Hardt2014}. Stakeholders not directly involved in decision-making are frequently excluded, resulting in their neglect in the optimization process \cite{Lopez2018, overdorf2018questioning}. Systems trained on specific domains struggle in new environments \cite{McMillan2011, Rodger2004, Sugiyama2017}. Optimization efforts tend to focus on benefiting high-priority users while disregarding others \cite{Huffaker2016, Tassi2016}. Risks from experimentation and parameter selection are often shifted onto users \cite{Bird2016}. Lastly, research in reinforcement learning also shows how optimization can amplify unintended outcomes \cite{amodei2016concrete, whittlestone2021societal, rathnam2024rethinking}. 

\vspace{0.15cm}

\noindent\textbf{Systems Dynamics.} Systems thinking (and theory) offer valuable frameworks for focusing on the study of interconnections and feedback loops \cite{meadows2008thinking, reader2022models}. This approach has been influential in fields such as organizational management, environmental sustainability, and policy design, where traditional reductionist methods often fall short \cite{sterman2000business}.  It has been applied across domains to address the challenges of unintended consequences, e.g., prior work shows the importance of systems thinking in decision-making under economic uncertainty, in predicting systems, and in identifying leverage points for sustainable solutions \cite{reader2022models, sterman2000business, meadows2008thinking, bertalanffy1968general}.

\section{Preliminaries and Definitions}\label{sec:prelim}

In this section, we explore three interconnected aspects of optimization: the mathematical foundations that define and solve problems (\S\ref{sec:mathopt}), the stakeholders and externalities that influence and are impacted by optimization (\S\ref{sec:stakeholders}), and the system's dynamics that govern the interactions and feedback mechanisms within complex systems (\S\ref{sec:systems}). 

\subsection{Mathematical Optimization}\label{sec:mathopt}

Mathematical optimization finds the best solution from feasible alternatives~\cite{Wang2023} and is vital in fields like economics, AI and ML, engineering, and operations research. The process involves defining the \emph{problem scope} that sets the boundaries (domain, temporal, and physical), specifying the \emph{variables} ($x$) and the \emph{objective function} ($f(x)$) that quantifies utility, determining the \emph{constraints} ($C$) for feasible solutions, identifying the corresponding \emph{solution space} ($X$), and selecting the \emph{evaluation metrics} ($M$) to assess solutions. The optimization problem is then formalized as:
\begin{align*}
\text{Optimize }_{x \in X} f(x)
\end{align*}
Once a solution is found, it is evaluated to verify its optimality, often requiring multiple iterations where previous solutions/evaluations inform refinements in subsequent steps.

\subsection{Stakeholders and Externalities}\label{sec:stakeholders}

Optimization, however, leads to unintended consequences. That is, in real-world optimization applications, there are generally agents or groups that are either directly interacting with or indirectly affected by optimization. These individuals are known as the {\emph stakeholders}~\cite{brauers2013optimization} and could be either direct participants of the optimization process (internal stakeholders) or those who are indirectly affected by the optimization (external stakeholders). Internal stakeholders often set objectives and constraints for the optimization and external stakeholders introduce additional considerations which are known as \emph{externalities} in economics \cite{pigou1920economics, coase1960problem}. Externalities can be positive (benefits to third parties) or negative (drawbacks to third parties) and are not reflected in the optimization's objective function.\footnote{Note that although certain stakeholders are considered internal from an organizational outlook, externalities can still impact them: The organization may fail to add their input into the optimization, preventing them from directly interacting with the optimization.}

Addressing externalities in optimization involves figuring out how to internalize them (i.e., quantify them), to ensure the optimization accounts for the impacts of externalities on the stakeholders~\cite{BEJAN202468}. Depending on the specific externality, this could involve different measures such as cost-benefit analysis (CBA)~\cite{Boardman2018} which is a comparison of costs and benefits of externalities and using them as an internalization quantitative framework, social welfare functions (SWF)~\cite{Arrow1951ARRIVA} which are an aggregation of utilities (and disutilities) from externalities, or a Cap-and-Trade System~\cite{investopediaTradeBasics}, which establishes a fixed limit on allowable externalities and enables the trading of permits to encourage compliance.\footnote{The quantification method used depends on the specific problem setting as well as stakeholders' goals in the optimization.}

In sum, this framework helps characterize and internalize unintended outcomes by discerning positive and negative types, external stakeholders, their goals, and relevant quantifications. However, no single economic method might fully capture the complexity of social cases. Other (even non-economic) methods might be more suitable for certain contexts. Even so, externalities offer valuable guidelines for analyzing stakeholders and the \emph{category} of required quantification methods. Despite their shortcomings, tools like CBA, SWF, and Cap-and-Trade still offer key practical utilities.

\subsection{Systems Dynamics}\label{sec:systems}

Externalities in complex socioeconomic systems are influenced by dynamic behaviors, feedback loops, and interconnections, which can amplify externalities over time \cite{Fleurbaey2023}. These interconnected dynamics make managing externalities challenging, especially in determining when and where to internalize them. While economic methods offer precise characterizations, they are too limited to fully address these complexities. Therefore, we suggest adopting \emph{systems thinking} as a broader analytical approach to gain deeper insights into these issues, as its normative view focuses on how systems should be designed and function, rather than simply describing their current state.

Drawing on systems thinking as a mindset, we also require a corresponding mathematical model to formally analyze systems, i.e., the ``theory." Systems theory provides a formal approach to examine how different parts of a system interact and influence each other to form a complex cohesive whole. Systems theory includes various frameworks, however, we refer to the layered architectures~\cite{sarjoughian2001,voros2005,jensen1970linear,Smith1973Hierarchical,Zhu2015HierarchicalAO,matni2024theory,empoweredFive}. They generally have established layers on different levels of abstraction where each layer addresses different aspects of the system's functioning and control. Layers organize components hierarchically and make understanding complex systems easier and more manageable by dividing them into specialized parts. This in turn helps us identify when and where to address externalities: lower levels focus on localized goals and higher levels analyze broader impacts. We use the following layered architecture:
\begin{enumerate}
    \item \textbf{Physical Layer}: This layer is about direct, tangible components of system (products, resources, infrastructure, or actions that are physically observable and measurable). It focuses on input-output dynamics, like resource allocation, production levels, or environmental changes. 

    \item \textbf{Regulatory Layer}: This layer sets rules, constraints, and enforcement mechanisms to govern behavior within the system. It introduces regulations or levies to influence entities and address externalities directly.

    \item \textbf{Supervisory Layer}: This layer ensures real-time monitoring, feedback, and adjustment of systems behavior. It uses physical and regulatory layers' data to dynamically optimize and correct deviations from desired outcomes.

    \item \textbf{Strategic Layer}: This layer encompasses the long-term planning and optimization of the system, integrating systemic goals, predictive modeling, and stakeholder objectives. It evaluates policies holistically, balancing competing priorities and externalities over time.
\end{enumerate}

\section{Systematic Externality Internalization }\label{sec:systems-theory}

We can now explore the use of the layered architecture from \S\ref{sec:systems} to analyze an optimization function, internalize externalities, examine the roles of various layers, and explain why externalities are addressed at a specific layer.

\subsection{Physical Layer: Operational and Direct System Decisions}

In the physical layer, the optimization considers internal stakeholders' goals and the externalities impacting external stakeholders. Let $G_{\text{int}}$ be the set of internal stakeholder goals, and $E_p^-(x), E_p^+(x)$ be the unexpected drawbacks and benefits at the physical layer. Let $f_{G_{\text{int}}}(x)$ capture the cost of the solution $x$ based on the goals $G_{\text{int}}$. Let $f_p(x) = f_{G_{\text{int}}}(x) + \gamma \cdot E_p^-(x) - \delta \cdot E_p^+(x) $ where $\gamma,\delta>0$ are penalty weights for externalities. The optimization can be written as, 
\begin{align*}
\min_{x} f_p(x) \;\; \text{  subject to  } \;\; x \text{  satisfies  } \{R(x) \le R_{\max},C_p(x)\}
\end{align*}
where $C_p(x)$ is the physical layer constraints, and the resources used $(R(x))$ satisfies $R(x) \le R_{\max}$, with $R_{\max}$ being the maximum available resources, which captures the resource limitations.

\vspace{0.12cm}

\noindent \textbf{Feedback Mechanism} The real-time adjustments would modify the solution based on the changes in resource availability, i.e., setting $x \leftarrow x-\eta \nabla R(x)$ where $\eta$ is a penalty weight and $\nabla R(x)$ is the gradient of $R(.)$ at $x$. Using the observed externalities, we would also readjust the parameters $\lambda,\gamma$, e.g., set $\gamma \leftarrow \gamma+\eta \frac{\partial E_p^-}{\partial x}$ where \(\eta>0\) is the adjustment factor based on stakeholder input.

\vspace{0.12cm}

\noindent \textbf{Why Internalize Externalities at the Physical Layer?} Externalities related to aspects such as resource consumption arise directly from actions in the physical layer. Internalizing them here ensures that the system accounts for these impacts in its decisions. The physical layer is an ideal location to internalize them as it directly interacts with them.

\subsection{Regulatory Layer: Compliance and Enforcement}

The regulatory layer would involve regulatory constraints captured by $C_r(x)$. Furthermore, the regulatory layer would involve taxation (using a tax function $T(x)$) and subsidies (using a subsidy function $S(x)$) to allow for adjusting the externalities. Let $f_r(x) = f_p(x) + T(x) - S(x)$. This results in the following optimization,
\begin{align*}
\min_{x} f_r(x) \;\; \text{  subject to } \;\; x \text{  satisfies } C_r(x)
\end{align*}
The objective at this layer depends on the physical layer.

\vspace{0.12cm}

\noindent \textbf{Feedback Mechanism}
This layer's feedback involves regulatory adjustments, i.e., continuous monitoring of compliance metrics and updating regulatory constraints ($C_r(x)$) as well as readjusting tax functions $T(x)$ and subsidy functions $S(x)$. The regulatory layer also involves policy feedback where stakeholders and supervisory data necessitate regulatory revisions to the constraints (\( C_r(x) \)).

\vspace{0.12cm}

\noindent \textbf{Why Internalize Externalities at the Regulatory Layer?} Regulations impose necessary constraints on externalities. Internalizing them at this layer ensures that the system complies with legal boundaries and respects societal standards. The regulatory layer enables the optimization to formally internalize externalities arising from regulatory oversight.

\subsection{Supervisory Layer: Monitoring and Real-Time Adjustments}

In this layer, we monitor system deviations. At each time point $t$, we consider the current solution $x$ and evaluate the solution using $M(x)$ (evaluation metrics in \ref{sec:mathopt}
) and compare against the target $M_{\text{target}}$. We also have dynamic constraints $C_s(x)$ that ensure the feasibility of solutions in real time. There are also positive externalities $E_s^+$, and negative externalities $E_s^-$, in this layer. Let $f_s(x) = \|M(x) - M_{\text{target}}\|^2 + \eta \cdot E_s^-(x) - \theta \cdot E_s^+(x)$ where $\eta,\theta$ are penalties for externalities. Utilizing $f_s(x)$, we get the objective,
\begin{align*}
\min f_s(x) \;\; \text{  subject to  } \;\; x \text{  satisfies  } C_s(x)
\end{align*}

\vspace{0.12cm}

\noindent \textbf{Feedback Mechanism} Here, feedback involves dynamic corrections, using supervisory data from monitoring systems and deviations from the targets to take corrective actions, i.e., update the solution such that $x \leftarrow x-\alpha \cdot \left(M(x) - M_{\text{target}}\right)^2$ with $\alpha$ being the learning rate. This layer also prompts externality adjustments, where weights are updated based on observed externalities using $\eta \leftarrow \eta + \beta \cdot \Delta E_s^-$ and $\theta \leftarrow \theta + \beta \cdot \Delta E_s^+$ where $\beta$ is an adjustment factor. Further, the layer gathers continuous feedback from stakeholders and adjusts decision variables or constraints in response.

\vspace{0.16cm}

\noindent \textbf{Why Internalize Externalities at the Supervisory Layer?} The Supervisory Layer plays a critical role in monitoring and refining system operations by evaluating performance and outcomes. Addressing externalities at this level enables the real-time identification and resolution of deviations and inefficiencies. By doing so, the system can maintain alignment with stakeholder goals without having to use higher levels to resolve deviations from the goals.

\subsection{Strategic Layer: Long-Term Planning and Sustainability}

In the strategic layer, we rely on long-term constraints captured by $C_l(x)$. We also evaluate the overarching system-level and long-term goals ($G_{\text{system}}$) such as sustainability, as well as internal and external stakeholder goals ($G_{\text{int}}$ and $G_{\text{ext}}$, respectively) and refine the goals as needed. Let $f_{G_{\text{system}}}(x)$ capture system goals' cost and $f_{G_{\text{int}} \cup G_{\text{ext}}}(x)$ be the cost of stakeholder goals at the strategic level. Let $f_l(x) =\alpha f_r(x) + \beta \cdot f_{G_{\text{system}}}(x)+ \gamma \cdot f_{G_{\text{int}} \cup G_{\text{ext}}}(x)$ where $\alpha,\beta,\gamma > 0$ are weights for balancing short-term and long-term objectives. In this layer, we get the following optimization,
\begin{align*}
\min f_l(x) \;\; \text{  subject to  } \;\; x \text{  satisfies  } C_l(x)
\end{align*}

\vspace{0.12cm}

\noindent \textbf{Feedback Mechanism} Here, feedback involves predictive updates where over periods of time, the deviations from expected outcomes are evaluated. This also involves modification of system and stakeholder goals, i.e., modifying system goals $G_{\text{system}}$ to include new long-term sustainability goals and the stakeholder goals $G_{\text{int}}$ and $G_{\text{ext}}$ to include the goals related to stakeholders with externalities observed.

\vspace{0.12cm}

\noindent \textbf{Why Internalize at the Strategic Layer?} Internalizing long-term externalities ensures that the system remains sustainable in the future. At the strategic level, the system interacts with high-level goals of optimization that affect its long-term behavior. Therefore, the goals that are intended to be long-term can be internalized in the strategic layer.

\begin{figure*}
    \centering
    \includegraphics[width=0.86\columnwidth]{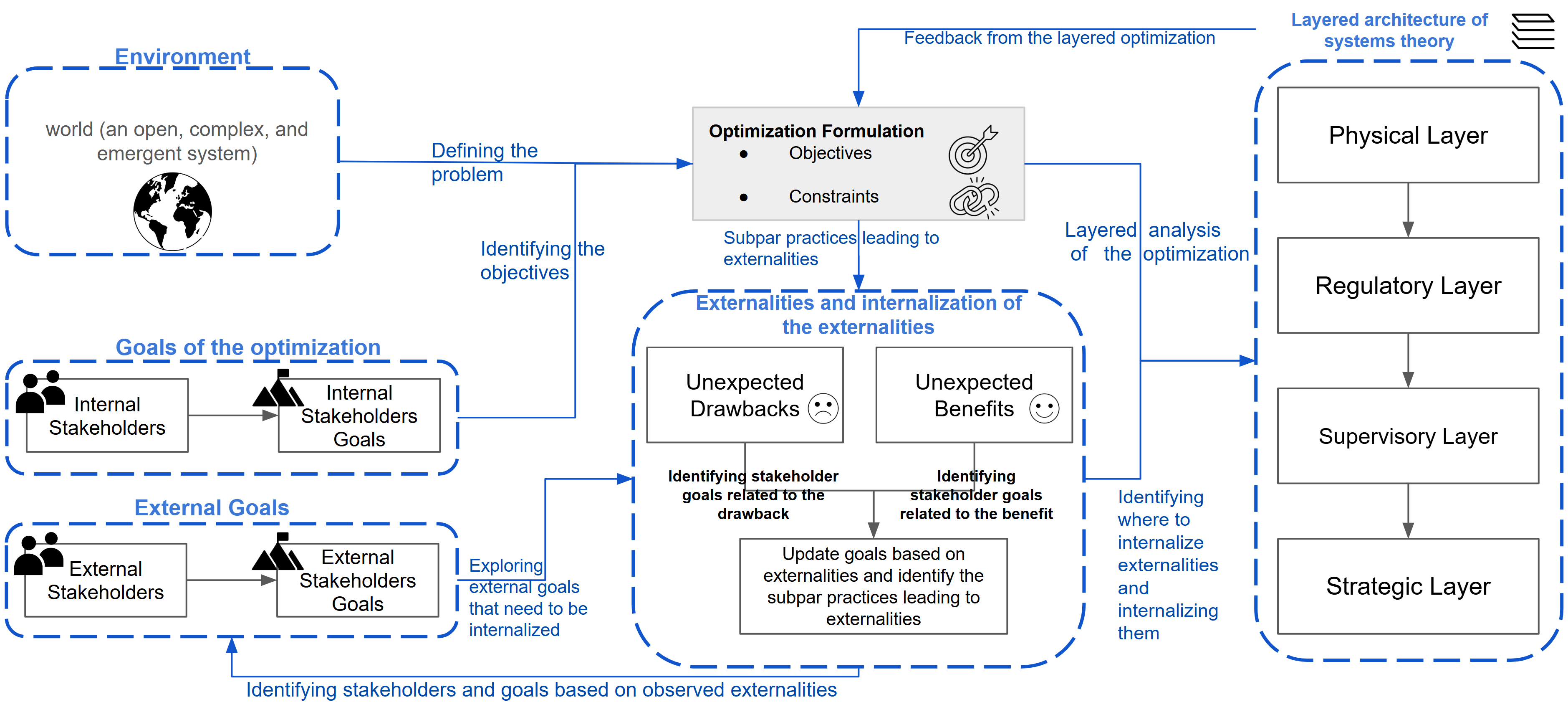}
    \caption{Framework overview: An optimization is defined with internal goals and the core task to optimize for. The optimization has externalities due to subpar practices and unmet external goals. Missed stakeholders, their goals, and the subpar practice are identified to revise goals and internalize externalities within the layered architecture. The optimization is then refined periodically by feedback.}
    \label{fig:flowchart2}
\end{figure*}
\section{Putting It All Together and Use Cases}

In sum, our framework for understanding and resolving unintended consequences is of the following steps:  
\begin{enumerate*}[label=(\arabic*)]    \item Begin with the original vanilla optimization, which focuses on objectives related to internal stakeholders, leading to unforeseen outcomes (descriptive).  
    
    \item Identify all affected stakeholders, including external stakeholders who are not directly involved in the optimization (and their goals) and utilize the economic framework of externalities to categorize the unforeseen outcome as either an unexpected drawback or benefit (descriptive). 
    \item Identify the cause of the externality, i.e., the subpar practice responsible, using the information from step 2 (descriptive).  
    \item Apply a pertinent quantification method, such as cost-benefit analysis (CBA), social welfare functions (SWF), Cap-and-Trade, or taxes, to internalize the externality based on its type (descriptive).  
    \item  Incorporate systems thinking (and the layered model) as an added outlook to address feedback and interconnections (normative). 
    \item Determine the most appropriate layer for internalizing externalities (normative).
\end{enumerate*} 
The logical overview for the entire framework is also depicted in Figure~\ref{fig:flowchart2}.

\vspace{0.12cm}

\noindent\textbf{Note on Identifying Subpar Practices.} Optimization involves various types of subpar practices. A notable advantage of using the economic framework of externalities is its ability to help us identify the specific subpar practice responsible for an externality. This is achieved by first identifying all stakeholders and their goals, as outlined in step 2 above.

We provide three examples to show how observable externalities relate to a specific subpar practice and how to resolve them using our proposed framework. The formulations presented are not the only possible choices but demonstrate how different subpar practices manifest in terms of externality internalization.  In our use cases, we identify the corresponding categories \cite{merton2003obituary, merton1936unanticipated} of subpar practices, as: \begin{enumerate*}
    \item \emph{ignorance} which refers to a lack of complete knowledge about the system, variables, or interactions in the optimization process. This often happens when the systems are complex and there is a lack of information to capture a complete picture, \item \emph{error} which occurs when incorrect assumptions, erroneous models, or incorrect reasoning lead to unintended outcomes. This can be caused by biases, oversights, or relying on incomplete or out-of-date information, and \item \emph{immediacy of interests} which stems from focusing too narrowly on immediate benefits while neglecting long-term impacts. Decisions made with short-term goals often disregard potential long-term effects or future externalities \cite{merton2003obituary, merton1936unanticipated}.

\end{enumerate*} We can now explore our use cases.

\begin{table*}
\small
\begin{tabular}{cp{.5\textwidth}p{.35\textwidth}}
\hline
Layer       & Optimization Function   & Feedback Loops   \\ 
\hline
\hline
Physical    & Let $T_j$ be the mean waiting time for an employee in time slot $t_j$. The congestion cost in this layer is defined as $L_{\text{physical}} = \sum_{j=1}^k T_j \cdot n_j$   &  Adjust the number of time slots $k$ using feedback, based on resource availability.  \\
\hline
Regulatory  & Ensuring scheduling feasibility: $\min \sum_{j=1}^k T_j \cdot n_j$ s.t. $\sum_{j=1}^k x_{ij} = 1$ $\forall \; i \in [N]$, $n_j \leq \Gamma_j$ and $x_{ij} \in \{0, 1\}$. $i$ is the employee index.& Use feedback to ensure regulatory compliance by checking if $n_j \le \Gamma_j$ and reassigning individuals to time slots where $n_j > \Gamma_j$. \\
\hline
Supervisory & Here, adjust policies to optimize productivity: $\min \sum_{j=1}^k T_j \cdot n_j + \sum_{i\ell} \text{CB}_{i\ell}(x)$
s.t. $\sum_{j=1}^k x_{ij} = 1$ $\forall \; i \in [N]$, $ n_j \leq \Gamma_j$, and $x_{ij} \in \{0, 1\}$ & Use feedback like employee-reported satisfaction and team outputs to modify $C_{i\ell}$ and $B_{i\ell}$. \\
\hline
Strategic   & Consider long-term goals like retention cost $R(x)$. The strategic layer aligns goals with satisfaction and productivity: $\min \sum_{j=1}^k T_j \cdot n_j + \sum_{i\ell} \text{CB}_{i\ell}(x) + R(x)$ s.t. $\sum_{j=1}^k x_{ij} = 1$, $n_j \leq \Gamma_j$ and $x_{ij} \in \{0, 1\}$& Periodic evaluation of retention rates/adjustment of long-term policies using supervisory insights.  \\ 
\hline
\end{tabular}
\caption{Layered Architecture Layout for Water Cooler Example}
\label{tbl:water-cooler}
\end{table*}

\subsection{Office Water Coolers and Social Interactions}

The water cooler effect, studied in social sciences~\cite{Weeks07Water, Orn22Covid, Lynne17Water}, explores how informal interactions impact employee productivity and well-being. This example explores a company implementing staggered time slots for water cooler access to enhance productivity while neglecting the water cooler effect and the importance of social interactions. We discuss the unexpected repercussions and how to integrate them into the optimization process to avoid such issues.

\vspace{0.12cm}

\noindent \textbf{Internal Stakeholders and Basic Optimization Formulation.}\label{sec:vanilla} To define the optimization formulation for the water cooler problem, we first need to identify the internal stakeholders of the optimization and their goals. Studies such as~\cite{Weeks07Water, Orn22Covid, Lynne17Water} investigating the importance of the water cooler effect and articles exploring the concept of stakeholders in a corporate context ~\cite{maddevs2024internal}, help us identify {\bf company leaders and management} as internal stakeholders. These stakeholders aim to increase productivity, reduce idle time, and ensure employees have sufficient off-task time.

Based on stakeholder goals, we can derive the following vanilla optimization problem.  For $N$ employees, assign them to $k$ time slots $T = \{t_1, t_2, \ldots, t_k\}$, each with a capacity $\Gamma_j$ for $j \in \{1,2,\dots,k\}$. Let $x_{ij}$ be a binary variable indicating if employee $i$ is assigned to time slot $t_j$, and $L_j$ be the loss function for delay cost at $t_j$. Let $n_j = \sum_{i=1}^N x_{ij}$ where $j \in \{1,2,\dots,k\}$. The company's objective is:
{\small
\begin{align}\label{eq:water-naive}
    &\min \sum_{j=1}^k L_j(n_j)\;\; \text{ subject to } \\
    &\sum_{j=1}^k x_{ij} = 1 \; \forall i \in \{1, 2, \ldots, N\}, \nonumber\\
    &n_j \leq \Gamma_j \; \forall j \in \{1, 2, \ldots, k\}, x_{ij} \in \{0, 1\}\nonumber
\end{align}}
where $\sum_{j=1}^k x_{ij} = 1 \; \forall i \in \{1, 2, \ldots, N\}$ implies each employee is assigned to exactly one timeslot and $n_j \leq \Gamma_j \; \forall j \in \{1, 2, \ldots, k\}$ implies that the number of people assigned to each time slot is within the capacity.

\vspace{0.12cm}

\noindent \textbf{External Stakeholders.}\label{sec:side-igno} We can now use the vanilla optimization formulation to investigate the outcomes of the optimization. The vanilla optimization focuses on minimizing the time wasted at the water cooler, aligning with internal stakeholders' primary goal. However, we need to expand our view to the external stakeholders as well. Referring to the literature on the water cooler behavior~\cite{Weeks07Water, Orn22Covid, Lynne17Water,maddevs2024internal}, we identify the following external stakeholders (and stakeholder goals): (1) {\bf Employees}, who desire opportunities for social interactions and positive relationships. Note that employees are external stakeholders because their goals were excluded from the optimized objectives, and they had no direct influence on the optimization process, and (2) {\bf Customers}, who seek effective services at an affordable cost.

\vspace{0.12cm}

\noindent \textbf{Unintended Consequences.} We can see that the vanilla optimization with the sole goal of reducing waiting time is unaware of the impact of the water cooler behavior on the employees or employees' and customers' external goals. Therefore, a solution derived from the vanilla formulation could result in {unexpected drawbacks}: \begin{enumerate*}
    \item Strict time slots reduce informal information and idea sharing, which are pivotal for building trust and a positive work culture~\cite{Lynne17Water, Orn22Covid}. \item This could harm interpersonal relationships, create tension, increase stress, and reduce performance, potentially leading to employee retention issues and a negative company image, \item All this could affect customer service quality as well as customer perception of the company. 
\end{enumerate*} Conversely, there are also {unintended benefits}, such as potentially enhanced employee privacy due to the reduction of informal interactions~\cite{Weeks07Water}.

\vspace{0.12cm}

\noindent \textbf{Why ignorance?} Given the set of unforeseen drawbacks and external goals, an intriguing question to explore is what causes these unexpected issues. In the water cooler example, the company aimed to reduce the time employees spent at the water cooler, with the sole objective of minimizing time. However, for the employees, social interactions were an essential aspect of their objectives. Because employees were excluded from the decision-making process, despite being directly affected by it, their goals were overlooked in the optimization. This disregard for employees' objectives and community dynamics led to unintended drawbacks. Thus, this example highlights the flawed practice of ignorance.

\vspace{0.165cm}

\noindent \textbf{Internalizing Externalities: A Systems Theory Framework.} With the externalities and their causes identified, the focus now is to make the optimization more resilient by internalizing these externalities. In the water cooler case, internalizing externalities requires incorporating social interactions. Since the goal is to improve productivity and reduce time spent on unproductive activities, we need to determine which social interactions contribute positively to productivity. We can consider the interactions between any two individuals, recognizing that these interactions come with costs, such as time taken away from work and logistical expenses needed to facilitate the interaction. At the same time, these interactions can bring benefits to the company, such as fostering idea generation. To incorporate social interactions, we would ideally need to ensure that benefits outweigh costs. One approach to achieve this is using cost-benefit analysis (CBA) to evaluate both the costs and benefits of interactions and adjust the objective by penalizing it per interaction costs in the solution. We can formally define the CBA-based internalization as: for individuals $i,\ell \in \{1,2,\dots,N\}$, let $C_{i\ell}$ be the cost and $B_{i\ell}$ be the benefit of their interactions. The cost-benefit for employees $i$ and $\ell$ is $\text{CB}_{i\ell}(x) = C_{i\ell}\cdot\mathbf{1}\left( \exists j \in \{1,2,\dots,k\} x_{ij}\neq x_{\ell j}\right)-B_{i\ell} \cdot \mathbf{1}\left( x_{ij}=x_{\ell j} \; \forall \; j \in \{1,2,\dots,k\}\right) $. This captures the interactions' benefits occurring in the same time slot and the cost of missing interactions when they are in different time slots. Using CBA, we internalize the externalities with $CB_{i\ell}$,
{\small
\begin{align*}
    &\min \sum_{j=1}^k L_j(n_j) + \sum_{i\ell} \text{CB}_{i\ell}(x) \text{ subject to }\\ &\sum_{j=1}^k x_{ij} = 1 \; \forall i \in \{1, 2, \ldots, N\}, \\
    &\sum_{i=1}^N x_{ij} \leq \Gamma_j \; \forall j \in \{1, 2, \ldots, k\}, x_{ij} \in \{0, 1\}
\end{align*}}
While CBA helps, it does not specify when internalization should occur during optimization. Thus, we use the layered architecture, which has a detailed view of connections and feedback in the process. A simple overview of the layered architecture for the water cooler example is in Table~\ref{tbl:water-cooler}. In this stack, the supervisory layer internalizes the loss of collaboration by considering the total cost-benefit, $\sum_{i\ell} \text{CB}_{i\ell}(x)$.

{\bf Remark.} The use of a supervisory layer to internalize externalities in this example is based on the fact that the externalities in the question stem from ignorance. To internalize them, we must be able to observe and track deviations from the desired outcomes 
(which were initially aimed at increasing productivity but shifted away from that goal due to fewer social interactions) and address the externalities caused by this ignorance leading to the deviation. The supervisory layer monitors deviations, making it ideal for incorporating ignorance-related externalities.

\subsection{Using AI to Select Candidates for Hiring}

With AI becoming a key tool in decision-making, its use in corporate recruitment has gained significant attention~\cite{gordon2023ai, ahmed2024navigating, reuters2018amazon, Chen2023}. AI recruitment relies on training data and parameters to decide which candidates move forward. This example considers a company using an AI tool based on historical data, with an assumption that the features derived from historical data represent diverse and best-fit candidates. We explore the unexpected optimization outcomes and how to address them.

\vspace{0.16cm}

\noindent \textbf{Internal Stakeholders and Basic Optimization.}\label{sec:vanilla-error} Our first goal is to formally define the basic vanilla optimization for the hiring problem. Before defining the formulation itself, we need to identify the internal stakeholders and their goals. Referring to prior research on AI in hiring~\cite{gordon2023ai, ahmed2024navigating, reuters2018amazon, Chen2023}, we identify the following internal stakeholders (and goals): (1) {\bf The organization, HR, and hiring managers} who aim to select the best candidates for interviews and hiring and to ensure fair representation in the candidate pool, eliminate discrimination, and avoid practices that may lead to litigation, and (2) {\bf Teams} who seek candidates that best fit their team.

Our objective is to select the best set of candidates given an AI trained on prior data. To formally define the optimization formulation, we first define a scoring function \( \text{Score}(x) = \sum_{i=1}^n w_i \cdot f_i(x) \), where \( f_i(x) \) represents feature $i$ of candidate $x$, and \( w_i \) represents weights derived from historical data. Let \( C \) represent the set of all candidates.  The scoring function calculates how well a candidate fits the observations from the training data. The goal is to select the top \( k \) candidates with the highest scores for interviews:

{\small
\[ \max_{X \subseteq C, |X| = k} \sum_{x \in X} \text{Score}(x).
\]
}

\vspace{0.12cm}

\noindent \textbf{External Stakeholders.}\label{sec:side-err} The goal of the vanilla optimization is to select top $k$ candidates that best fit the selection criteria based on past data. However, noting prior work on AI hiring~\cite{gordon2023ai, ahmed2024navigating, reuters2018amazon, Chen2023}, we identify the following external stakeholders (and goals): (1) {\bf Candidates} who want to ensure fair and equal opportunities for selection, and (2) {\bf Regulatory agencies} who want to ensure fairness and bias-free hiring practices.

\vspace{0.12cm}

\noindent \textbf{Unintended Consequences.} The unexpected drawbacks affecting external stakeholders are: \begin{enumerate*}
    \item AI tools may introduce bias based on factors like gender, age, and disabilities~\cite{reuters2018amazon, ahmed2024navigating}. \item AI could select unqualified candidates over more qualified individuals~\cite{ahmed2024navigating}, resulting in perceived biases and potential litigation. \item  Consequently, a lack of diversity in selected candidates could hinder innovation in the organization.
\end{enumerate*} 

\begin{table*}
\small
\begin{tabular}
{cp{.5\textwidth}p{.35\textwidth}}
\hline
Layer       & Optimization Function   & Feedback Loops\\ 
\hline
\hline
Physical    & 
The AI assigns a score \( \text{Score}(x) = \sum_{i=1}^n w_i \cdot f_i(x) \) to each candidate, where \( f_i(x) \) is the value of feature \( i \) for candidate \( x \), and \( w_i \) is the weight for feature \( i \) derived from historical hiring data. Let \( C \) represent the set of all candidates. The objective is to select a subset \( X \) of size \( k \) such that: \( \max_{X \subseteq C, |X| = k} \sum_{x \in X} \text{Score}(x) \).
&  The weights $w_i$ are updated after analyzing the set of selected candidates.\\
\hline
Regulatory  & The lack of diversity is mitigated. Here, we introduce Cap-and-Trade bounds, leading to optimization \( \max_{X \subseteq C, |X| = k} \sum_{x \in X} \text{Score}(x)\) s.t. $h_i(X) \ge h_{\text{threshold}}(X) \; \forall i \in \{1,2,\dots,k\}$. & Feedback should be used to adjust the $h_{\text{threshold}}(X)$. \\
\hline
Supervisory & Monitor the diversity and optimality of the solution. Let \( \text{Div}(X) \) be the diversity of solution \( X \) and $\text{Div}_{\text{target}}$ be the target diversity, then in this layer, the optimizer evaluates $\|\text{Div}(X)-\text{Div}_{\text{target}}\|$ to calculate deviations from the desired diversity. & Feedback entails modifying solution $X$ based on the deviation from $\text{Div}_{\text{target}}$ according to organizational and regulatory requirements.\\
\hline
Strategic   & Optimizes hiring for long-term goals like diversity ($G_{\text{Div}}$), creativity ($G_{\text{Create}}$), and retention ($G_{\text{Reten}}$): $G = \omega_1 \cdot G_{\text{Div}} + \omega_2 \cdot G_{\text{Create}} + \omega_3 \cdot G_{\text{Reten}}$, with  $\omega_1, \omega_2, \omega_3$ representing the weights for each goal.
  & Evaluate the alignment between hiring outcomes and stakeholder goals by using and adjusting $\omega_1,\omega_2,\omega_3$ or objectives based on the deviations.  \\ 
\hline
\end{tabular}
\caption{Layered Architecture Layout for AI Hiring Example}
\label{tbl:AI-Hiring}
\end{table*}

\vspace{0.12cm}

\noindent \textbf{Why Error?} Here, the optimization assumed that past training data accurately represented diverse subgroups, apt candidates, and was free from biases. However, this assumption was flawed, as past data may contain subtle biases that AI tools can amplify with the training, leading to biased results. A notable example is Amazon's automated recruitment system, which was found to be biased against female applicants~\cite{reuters2018amazon}. Despite the company's claims of commitment to equal representations, erroneous data assumptions led to unexpected drawbacks in the optimization, aligning with the ``error" category in subpar practices.

\vspace{0.16cm}

\noindent \textbf{Internalizing Externalities: A Systems Theory Framework.} Having identified the externalities arising from the vanilla optimization, the next step is to internalize them. In this case, the optimization produces biased solutions due to a flawed assumption. To address this, a method is needed to ensure the optimization avoids such biases. We propose the use of Cap-and-Trade systems. A Cap-and-Trade system enables the optimizer to introduce a threshold on the diversity required in the solution. This is motivated by rules such as the EEOC 4/5 rule which flags potential bias if a group's hiring rate is under 80\% of the highest group's rate~\cite{eeoc1978uniform}. In this use case, we use the Cap-and-Trade system to meet a specific threshold of diversity in hiring decisions to comply with the 4/5 rule. This ``cap" ensures that hiring practices do not disproportionately disadvantage protected groups. We use a lower-bound cap and no trading. Assume we have a set of $k$ groups with hiring rates $h_1(X),h_2(X),\dots,h_k(X)$ for the solution $X$, and a minimum threshold hiring rate $h_{\text{threshold}}(X)$, e.g., $h_{\text{threshold}}(X)=0.8\max_{i} h_i(X)$.  Then our objective is: 

{\small
\begin{align*}
&\max_{X \subseteq C, |X| = k} \sum_{x \in X} \text{Score}(x) \text{ subject to }\\ &h_i(X) \ge h_{\text{threshold}}(X) \; \forall i \in \{1,2,\dots,k\}
\end{align*}}

To determine when internalization occurs, like before, we use the layered architecture. A simple overview of the layered architecture for the AI hiring example is provided in Table~\ref{tbl:AI-Hiring}. In this use case, the regulatory layer internalizes diversity needs by incorporating Cap-and-Trade bounds.

{\bf Remark.}  The use of this layer to internalize error-related externalities is grounded in the understanding that errors arise from factors like flawed assumptions and biases. The regulatory layer is responsible for managing regulations and ensuring solutions comply with accepted social standards and legal boundaries. For instance, adherence to regulations like the EEOC 4/5 rule should be managed at this layer.

\subsection{Aggressive Campaigns for Quarterly Sales}

Aggressive marketing campaigns are commonly used to maximize short-term profits, despite potential long-term consequences~\cite{Kai14Harm, Hanssens04Prom}. This example explores a company using aggressive marketing to boost quarterly sales. We examine the unexpected outcomes of this scenario and how to incorporate them into the optimization.

\vspace{0.16cm}

\noindent \textbf{Internal Stakeholders and Basic Optimization.}\label{sec:vanilla-immediate}
We first need to identify the internal stakeholders and their goals to define the vanilla optimization. Referring to the literature on aggressive marketing campaigns~\cite{Kai14Harm, Hanssens04Prom, Kessler2024, Lehman97Long, Mela98Long}, we identify internal stakeholders as the {\bf company leaders} who aim to increase quarterly profits. These stakeholders want to increase quarterly sales through aggressive marketing and discounting. Thus, the basic optimization focuses on maximizing short-term sales by allocating marketing budget and discounts efficiently, subject to operational and budget constraints. Formally, let $x$ represent the decision variables, where $x = {m_1,m_2,\dots,r_1,r_2,\dots}$ denotes marketing cost and discount rates for product $i$. The optimization problem is,
{\small
\begin{align*}
&\max_{x} \; \sum_{i=1}^{n} p_i(x) \cdot s_i(x) \text{ subject to }\\
&\left \{\sum_{i=1}^{n} m_i(x) \leq B, s_i(x) \le \text{Inventory}_i \; \forall i, p_i(x) \geq p_{\text{min}} \; \forall i\right\}
\end{align*}}
where $p_i(x)$ is the product price, $s_i(x)$ is the sales volume, $\text{Inventory}_i$ is the inventory size, $p_{\text{min}}$ is the minimum price, and $B$ is the marketing budget.

\paragraph{External Stakeholders.}\label{sec:side-imme}

Using prior work on aggressive campaigns~\cite{Kai14Harm, Hanssens04Prom, Kessler2024, Lehman97Long, Mela98Long}, we note external stakeholders (and goals) as: (1) {\bf Customers} who seek quality items at reasonable prices and desire long-term reliability from the company and its products,  (2) {\bf Competitors} who seek to maintain a competitive market and prevent market monopolization by the company.

\vspace{0.12cm}

\noindent \textbf{Unintended Consequences.} Maximizing short-term profits while ignoring long-term impacts may result in unexpected drawbacks for external stakeholders: \begin{enumerate*}
    \item Aggressive marketing encourages discount-seeking and stockpiling behavior~\cite{Mela98Long}, while reducing regular consumption in anticipation of future promotions~\cite{Kessler2024}. \item It may obscure product issues and their societal impact, especially in sensitive industries like healthcare~\cite{Kai14Harm}. \item Aggressive campaigns can also lead to market instability and unhealthy competition.
\end{enumerate*}  Conversely, aggressive campaigns do not always have drawbacks. With quality products, well-planned marketing can foster customer loyalty (unexpected benefit)~\cite{Kessler2024}.

\vspace{0.12cm}

\noindent \textbf{Why Immediacy of Interests?} Aggressive campaigns to boost profits inherently prioritize short-term gains over long-term sustainability. These campaigns focus on immediate objectives, often neglecting long-term profitability and sustainability. This, by definition, makes them a clear example of subpar practices driven by the immediacy of interests.

\begin{table*}
\small
\begin{tabular}
{cp{.5\textwidth}p{.35\textwidth}}
\hline
Layer       & Optimization Function   & Feedback Loops   \\ 
\hline
\hline
Physical    & $\max_x \sum_{i=1}^n p_i(x) \cdot s_i(x)$ subjected to $s_i(x) \le \text{Inventory}_i \; \forall i$ and $\sum_{i=1}^{n} m_i(x) \leq B$ where physical constraints, including inventory limits ($\text{Inventory}_i$) and a budget constraint ($B$), are incorporated to ensure feasible solutions while maximizing the objective.
&  Adjust inventory and budget allocations based on the solution. \\
\hline
Regulatory  &
$\max_x \sum_{i=1}^n p_i(x) \cdot s_i(x)$ s.t. $s_i(x) \le \text{Inventory}_i \; \forall i$, $ \sum_{i=1}^{n} m_i(x) \leq B$, and $ p_i(x) \ge p_{\min} \; \forall i$ where regulatory constraints are introduced to prevent aggressive low pricing that could lead to price gouging.  & Adjust the price thresholds by updating $p_{\min}$ as follows: $p_{\min} \leftarrow p_{\min}+\eta\sum_{i=1}^n \frac{\partial p_i(x)}{\partial x}$ where $\eta$ is a learning rate. \\
\hline
Supervisory & Evaluate solutions by balancing profit and cost through $\max_x \sum_{i=1}^n \big(p_i(x) \cdot s_i(x) - m_i(x)\big)$, subjected to $s_i(x) \leq \text{Inventory}_i \; \forall i$, $\sum_{i=1}^n m_i(x) \leq B$, and $p_i(x) \geq p_{\min} ; \forall i$. & Feedback uses sales data to assess optimization outcomes and adjust solutions to address deviations. \\
\hline
Strategic   & Maximize the SWF: $W(x)$.
  & Feedback updates $U_D(x), U_c(x)$ and $U_u(x)$ using long-term observations.  \\ 
\hline
\end{tabular}
\caption{Layered Architecture Layout for Aggressive Marketing and Promotions Example}
\label{tbl:aggressive}
\end{table*}

\vspace{0.15cm}

\noindent \textbf{Internalizing Externalities: A Systems Theory Framework.} To internalize externalities from the immediacy of interests, we suggest a social welfare function $W(x)$. SWF is based on the utilities gained by internal and external parties and the disutility from externalities. These utilities and disutilities could be aggregated over time. Thus, SWF allows heeding the long-term effects on the company, consumers, and market, and reducing the impact of short-term interests.  

Let $U_c(x)=\sum_{i=1}^n \left(p_i(x)\cdot s_i(x)-m_i(x) \right)$ represent the company’s utility, $U_u(x)=\sum_{i=1}^n \left(v_i(x)-p_i(x)\right)$ (where $v_i(x)$ is the intrinsic value of the product $i$ in regards to the consumer utility), and $U_D(x)$ be the disutility from externalities. The SWF is defined as $W(x) = U_c(x) + U_u(x) - U_D(x)$. Then the optimization problem is,


{\small
\begin{align*} 
&\max_{x} \; \sum_{i=1}^{n} p_i(x) \cdot s_i(x) + \lambda W(x)\text{ subject to }\\&\left \{\sum_{i=1}^{n} m_i(x) \leq B, s_i(x) \le \text{Inventory}_i \; \forall i, p_i(x) \geq p_{\text{min}} \; \forall i\right\} 
\end{align*} }
Here, $\lambda$ is a weight parameter. A simple overview of the layered architecture for aggressive marketing is listed in Table~\ref{tbl:aggressive}. In this example, the strategic layer internalizes externalities using SWF, which allows evaluation of short-term and long-term goals based on the loss of utility in SWF.

 {\bf Remark.} This layer is suited to internalize \textit{immediacy of interest} externalities due to its long-term planning to balance short-term and long-term gains and ensuring short-term goals do not compromise a company's long-term success.

\section{Discussions and Limitations}

In sum, we examined how to better characterize unintended consequences in optimization, identify all impacted stakeholders, and determine when and how to internalize externalities. We proposed integrating externalities with systems thinking to develop a comprehensive framework for navigating these challenges in complex, interconnected systems.



\subsection{Normative Directions and Value Conflicts}

Externalities can present value trade-offs. So we first discuss some points on research on values in AI that cover three aspects: embedding value analysis through \textit{Value Sensitive Design (VSD)} \cite{friedman1996value,friedman2002value,friedman2019value, umbrello2021mapping, sadek2023designing, hopkins2023value}, \textit{managing stakeholder tensions} in sociotechnical systems \cite{kallina2024stakeholder, heymans2024identifying, katsikeas2023artificial}, and addressing \textit{trade-offs in value conflicts} like fairness versus efficiency \cite{victor2024medical,fan2024minion, microsoft2025best}. Together, these strands emphasize the need for designs integrating values, stakeholder outlooks, and normative trade-off reasoning into optimization. E.g., \textit{how do we decide whether to trade a one-hour loss of road construction efficiency over the displacement of a household?} 

We recognize that such value conflicts demand normative judgment. Our framework, however, does not aim to resolve these tensions outright. Instead, it seeks to ensure they are acknowledged and methodically incorporated into the optimization process, rather than being overlooked. Also, our approach may not offer definitive resolutions, but externality quantification tools like Cost-Benefit Analysis help surface and partially (and structurally) navigate trade-offs. In detail:

\vspace{0.12cm}

 \noindent \textbf{The Normative Dimension is Central.}  Systems thinking may not resolve conflicts, but 1) it reframes optimization problems beyond efficiency maximization by broadening what is considered relevant in optimization. This reframing is inherently normative, 2) it normatively asks where and how externalities must be incorporated into the optimization process. Even if a choice is made to prioritize household displacement over efficiency, a key question remains: at which level of abstraction should this externality be addressed? Systems thinking navigates this normative decision.

\vspace{0.12cm}

 \noindent \textbf{Recent Normative Research.} Thus, our work aligns directly with recent normative conversations. Key papers from recent years on responsible AI, AI governance, and algorithmic fairness like \cite{chan2023harms, hutchinson2022evaluation, donia2022normative, laufer2023optimization} examine normative directions in algorithmic, ML, and optimization decision settings to limit harms. Our work contributes to these debates by structuring externalities, feedback loops, and value commitments in optimization.

\vspace{0.154cm}

 \noindent \textbf{Value Conflicts, Traditional Stakeholder Analysis, and Value-Sensitive Design (VSD).} Our framework does not ignore trade-offs but explicitly makes them visible/actionable within a broader system. It extends VSD and stakeholder tension frameworks by adding systemic and procedural dimensions often missing in VSD and stakeholder analyses. It offers a formal lens on systemic impacts that stakeholder analysis often overlooks: Externalities are unintended, system-wide consequences of pursuing specific goals, often invisible in stakeholder frameworks when actors cannot foresee or articulate them. Unlike traditional stakeholder analysis focused on direct relationships, externalities with systems thinking emphasize second and third-order system effects, including long-term social and institutional spillovers. In AI/ML, focusing solely on stakeholder tensions captures direct harms but misses broader effects harming groups not initially framed as stakeholders. 
 
 Our work (and math) complement VSD by: 1) focusing on systemic/indirect externalities beyond stakeholder engagement, 2) embedding externality analysis within optimization itself, not as an afterthought. These points show another \textit{extra} aspect of our contributions pertaining to these lines of work. Thus, since our proposition does not replace VSD but complements it,  comparisons should be made with vanilla optimizations without any systematic externality considerations (not with VSD or trade-off analyses that work alongside our framework, not in competition with it).

No framework can perfectly resolve deep value conflicts. It matters, however, how well frameworks: 1) surface critical questions and tensions, 2) ensure recognition of externalities and indirect harms, and 3) prevent optimization with narrow goals. The six-step method does all of these. It: 1) makes externalities explicit (Steps 1–4), 2) facilitates strategic and regulatory engagement (Steps 4-6), and 3) expands the optimization scope to include systemic consequences.

\subsection{Limitations}

First, subpar practices can also stem from data collection and sampling (not discussed in our work), leading to unmet goals \cite{dorfler2024towards}. Also, our framework helps identify causes such as error or ignorance, but comprehensively identifying all causes, stakeholders, and their goals remains a challenge. Externalities provide structure but not a full methodology for this. Moreover, categorizing issues (ignorance, error, or short-term focus) is context-dependent; e.g., a single scenario may be interpreted differently based on organizational priorities. Opting for parameters (like objective weights, short- vs long-term goals, or feedback terms) also depends on context. Lastly, tools like CBA or SWF help quantify externalities, but they may sometimes serve better as guiding perspectives and require adaptations to fit specific socioeconomic contexts. These decisions should be informed by stakeholder goals and require expertise to align with individual, organizational, and societal resolutions.

\bibliographystyle{unsrt}  
\bibliography{references}

\end{document}